\documentclass{article}

\usepackage{microtype}
\usepackage{graphicx}
\usepackage{subcaption}
\usepackage{booktabs}
\usepackage{hyperref}
\usepackage{amsmath,amssymb}
\usepackage[capitalize,noabbrev]{cleveref}

\usepackage{algorithm}
\usepackage{algorithmic}

\usepackage[accepted]{icml2026}
\usepackage{xcolor}

\makeatletter
\renewcommand{\Notice@String}{Accepted at the AI for Science workshop (ICML 2026).}
\makeatother

\icmltitlerunning{Adversarial Fast-Moving Real-World Domains for Evaluating AI Scientists}

\begin{document}

\twocolumn[
\icmltitle{Adversarial Fast-Moving Real-World Domains \\as Test Beds for Benchmarking AI Scientist Capabilities}

\begin{icmlauthorlist}
\icmlauthor{William Bolton}{aff1}
\icmlauthor{Philip Torr}{aff1}
\end{icmlauthorlist}

\icmlaffiliation{aff1}{Department of Engineering Science, University of Oxford, Oxford, United Kingdom}
\icmlcorrespondingauthor{William Bolton}{will.bolton@eng.ox.ac.uk}
\icmlkeywords{AI scientists, benchmarking, large language models, reasoning, novelty, hypothesis formulation, adversarial real-world domains, time-delayed ground truth}

\vskip 0.3in
]

\printAffiliationsAndNotice{}

\begin{abstract}
Benchmarking the ability of AI scientists to generate novel ideas is notoriously difficult. Existing benchmarks in this field have made progress in evaluating scientific reasoning and research replication, but often rely on synthetic tasks or retrospective targets, which may be confounded by prior exposure. We hypothesize that complex, adversarial, fast-moving real-world domains where expert practitioners independently generate observable outputs can provide a practical solution to fill this gap and evaluate the capabilities needed for AI scientists, including reasoning, novelty, and hypothesis formulation. We instantiate this framework in two structurally different domains, Formula 1 (F1), where models ideate around car design concepts for the 2026 season, and real pre-season innovations provide a ground truth, and Magic: The Gathering (MTG), where models propose decks from a recently updated card pool and are evaluated against 19 Pro Tour (PT) decklists. Across both domains, models produce plausible outputs, but few align with real-world expert solutions. In F1, the best model, GPT-5.2 matched 10 of 40 real innovations with 166 ideas proposed across runs. In MTG, the best deck from Gemini 3 Flash recovered 5 of 7 new-set cards from the third-place PT deck, and across all 108 decks, the cards models selected most often were also the cards most widely adopted by PT decks (Spearman $\rho = 0.74$, $p = 0.0003$). These results suggest that a key capability gap for AI scientists is not idea generation, but filtering, prioritization, and coherent novelty.

\end{abstract}

\section{Introduction}

AI is increasingly being used to augment scientific discovery. Advances in reasoning and test-time compute have made models capable knowledge workers \cite{snell2025scaling}. However, unlike the rapid advancement seen in coding agents, automatic verification of science is often difficult and slow \cite{cornelio2025need}. Furthermore, the ability of models to be truly novel is a key requirement if we are to realize the vision of AI scientists, autonomously progressing the frontier of research \cite{kitano2021nobel, lu2024ai, zhang2025exploring}. So how can we benchmark discovery systems when we don't know the answer?

We propose that real-world, adversarial domains where outcomes are observed after a time delay could act as useful benchmarks for AI scientists' capabilities. By adversarial, we mean there is some degree of competition between experts, which drives human innovation. Being in public ensures it can benefit the open source community, and after a period of time, we can verify whether model outputs are genuinely novel or consistent with human intuition.

\paragraph{Contributions.}
\begin{itemize}
    \item We propose a framework for evaluating AI scientists' capabilities with adversarial fast-moving real-world domains.
    \item We instantiate a proof of concept of the framework in two structurally distinct domains, Formula 1 car design under new technical regulations and Magic: The Gathering deck construction under evolving cards.
    \item We identify failure modes that are informative for AI scientist evaluation, including the production of plausible-but-misaligned ideas, and systematic blind spots for value that arises from complex surrounding contexts.
\end{itemize}

\newpage
\section{Related Work}\label{sec:related}

\subsection{AI scientists and hypothesis generation}

There is significant interest in the potential for agentic systems to autonomously accelerate science. Recent industrial efforts have shown comprehensive pipelines that ideate, experiment, and write up research with limited human supervision \citep{gottweis2025towards, lu2024ai, mitchener2025kosmos}. For hypothesis generation, prior studies have optimized for novelty against a literature corpus \citep{wang2024scimon} and compared large language model (LLM)-generated and human research proposals \citep{si2024can}.

\subsection{Benchmarking hypothesis and scientific reasoning}
Several recent benchmarks have made progress in evaluating scientific reasoning, hypothesis generation, and research replication. DiscoveryBench frames data-driven discovery as a structured hypothesis search over curated datasets~\citep{majumder2024discoverybench}, HypoBench provides a principled setup for hypothesis-generation evaluation~\citep{liu2025hypobench}, PaperBench measures whether agents can replicate published machine-learning papers and their code from the paper content alone~\citep{starace2025paperbench}, and ScienceAgentBench tests agents on curated data-driven science tasks~\citep{chen2024scienceagentbench}. These benchmarks advanced the field but largely rely on decomposed tasks, retrospective targets, or rubricable outputs. They therefore do not directly test prospective novelty in open-ended real-world domains. They are also susceptible to data contamination, since their ground-truth artifacts predate current model training cutoffs.

\subsection{Temporal evaluation}

One interesting research direction is using history as a test bed. A model could be trained with information up to a particular point in time and asked to recover subsequent discoveries. This idea has been explored in concept and in early implementations~\citep{goettlichetal2025, pittphilsci28024}, but to our knowledge not rigorously instantiated at scale. Historical replay is appealing, but clear pitfalls exist as clean knowledge cutoffs are hard to define, experiments are expensive, and a negative result is ambiguous because it may reflect limited capability, an insufficient search budget, or a mismatch with the discovery path humans actually took. Prospective evaluation negates many of these challenges. DeepMind demonstrated that once given to researchers, their co-scientist could help develop new hypotheses and research proposals \citep{gottweis2025towards}. This is potentially the best type of evaluation possible, but it is not accessible to all researchers and lacks a clear counterfactual. Potentially the closest prior work to this research evaluates models on forecasting tasks. Here, systems predict the resolution of pre-registered real-world questions whose answers are revealed only after a cutoff ~\citep{karger2024forecastbench, halawi2024approaching}. Forecasting and the time-delayed artifact matching we propose share the structure of reasoning under post-cutoff uncertainty, but differ in their target. Forecasting evaluates probabilistic prediction over predefined questions, whereas our framework evaluates open-ended ideation against an emerging real-world expert reference set.

\section{Evaluation Framework}\label{sec:framework}

Our framework evaluates AI scientist capabilities through a time-delayed generation task. For each domain, we define an information cutoff $t$ and construct a fixed input corpus containing only material available before that point, such as rules, constraints, public context, and relevant prior examples. The AI system is then asked to generate structured candidate artifacts from this corpus, such as ideas, designs, or solutions. These generations are evaluated against post-cutoff artifacts produced by real-world experts. This  creates a prospective-style evaluation of whether models can generate non-trivial ideas under realistic constraints, while removing the risk of information leakage.

A domain is suitable for this framework if it satisfies four criteria. First, it must have a constrained but non-trivial design space, such that outputs can be judged against explicit rules or objectives. Second, it must be adversarial such that it involves genuine expert innovation, rather than simple retrieval or optimization against a known answer. Third, it must produce publicly accessible observable downstream artifacts that can serve as the time-delayed ground truth. Fourth, models and the input context must have a clear information cutoff, so that models can be evaluated using only information plausibly available before the relevant innovations became public.

Each domain is treated as a fixed-corpus ideation problem where structured candidates such as ideas or solutions are produced, and evaluation is conducted along two axes. First, outcome similarity measures whether a generated artifact corresponds to an independently observed expert innovation in the same domain under the same constraints. This tests if the AI system can recover new ideas that later prove useful or relevant in the real world. Second, intrinsic quality measures whether the idea is well-grounded, rule-compliant, and plausible.

\subsection{Experimental Domains}\label{sec:domains}
Two independent domains were chosen to instantiate this framework: F1 2026 regulations, where free-form car design ideas are generated, and Magic: The Gathering (MTG) Lorwyn Eclipsed, a non-trivial but discrete card selection task. Both domains met all criteria, with corpus information released after the training cutoff of all models in the study.

\subsubsection{F1 2026 Regulations}

F1 is the pinnacle of motorsport and is fiercely innovative. Most often, the fastest car wins, and hence there is significant engineering investment from teams to develop new ideas that shave tenths of a second off lap times. In 2026, the FIA Formula~1 Technical Regulations went through the largest change in history. Teams have had to design new cars from the ground up to comply with these regulations. The corpus contained a 264-page PDF document FIA 2026 \emph{Section~C} Technical Regulations, dated 10th of December 2025.  The task was to produce technical innovations specific to the 2026 cars. Ground truth was a curated set of 40 real F1 concepts drawn from publicly available pre-season analysis (Appendix \autoref{tab:f1-sources}).

\subsubsection{MTG Lorwyn Eclipsed}

MTG is a competitive collectible card game in which professional players invest hundreds of hours per set identifying which new cards are most competitive. Lorwyn Eclipsed is an MTG set released on 23 January 2026, containing 267 unique cards. PT Lorwyn Eclipsed, the first competitive event using the new set, was held between 30 January and 1 February 2026. The corpus input combined the 267 new cards with 4{,}168 existing Standard-legal cards. The task was to produce a complete tournament-legal deck, prioritizing the new Lorwyn Eclipsed cards considered most viable. Ground-truth decklists were collected for the top 15 PT decks alongside four featured builds (Appendix \autoref{tab:mtg-sources}).

\section{Experimental Setup}\label{sec:setup}

\subsection{Three-agent pipeline}

A 3-agent pipeline was used in both domains to decouple three core skills of strategic intent, technical analysis, and generation. For F1, agent~1 receives a 2026-context summary and produces performance goals; agent~2 receives the full technical regulations and produces a regulatory mapping that tags each \emph{degree of freedom} with one of five loophole categories (\texttt{INTERFACE\_COUPLING}, \texttt{DEFINITION\_EDGE\_CASE}, \texttt{BROAD\_FREEDOM}, \texttt{NOT\_PROHIBITED}, \texttt{EXCEPTION\_ZONE}) and a list of cited article numbers; finally, agent~3 receives the outputs from agents 1 and 2, regulatory text for every article cited, and few-shot examples from F1 history, and produces approximately 15 car design ideas. Each generated idea inherits the union of loophole types from those mappings whose cited articles overlap its enabling articles; the full tagging procedure and verdict scales are detailed in the \hyperref[loophole-tagging]{supplementary methods}. For MTG, agent~1 receives the full 267 Lorwyn Eclipsed cards and labels each as \texttt{STAPLE}, \texttt{ROLE\_PLAYER}, \texttt{BUILD\_AROUND}, or \texttt{UNPLAYABLE}; agent~2 receives agent~1's shortlist together with condensed text for the pre-existing cards and pre-tournament metagame context, and produces a combined card shortlist with strategy notes; finally, agent~3 receives agent~2's output and generates three decks, each representing a different strategic angle. All outputs use structured JSON. For both domains, the pipeline was run in two configurations. In F1, the general configuration produced outputs across the whole car, while by-component narrowed the generation scope to specific car areas, such as the floor or rear wing. This tested whether reducing the search space improved output quality. For MTG, in one-shot mode, agent~3 outputs a single structured-JSON response, while in tool-use mode, it builds each deck via a tool-calling loop. This isolates the effect of construction-time scaffolding on the generated artifacts. The pipelines for each domain were run in both configurations against six frontier LLMs (GPT-5.2, o3, Gemini 3 Flash, Gemini 3.1 Pro, Qwen3 235B, and Qwen3 235B-thinking) chosen because their reported knowledge cutoffs predate both evaluation corpora and because they support long context windows.

\subsection{Evaluation metrics}

Novelty matching in the F1 domain utilized three stages. First, generated ideas and real innovations were embedded and shortlisted using a dual top-$k$ retrieval strategy ($k=3$), retaining the nearest real innovations for each generated idea and the nearest generated ideas for each real innovation. Shortlisted pairs were assessed by an LLM judge (GPT-5.5) for \texttt{MATCH}, \texttt{PARTIAL}, or \texttt{NO\_MATCH}. Candidate matches then underwent human review to verify if the idea described a similar physical effect to the real innovation. In parallel, each generated idea was assessed along three quality dimensions: citation accuracy, rule compliance, and engineering plausibility. An idea was treated as passing these quality filters only if its cited regulations were substantively accurate, its design was legal or plausibly within a regulatory grey zone, and its mechanism was physically plausible.

MTG evaluation used deterministic decklist comparisons rather than semantic judging. For each generated deck and ground-truth PT deck pair, we computed card overlap and similarity metrics (Jaccard, precision, and recall) for all non-land cards, and most relevant to the prospective nature of this framework the new non-land Lorwyn Eclipsed cards. Legality was assessed by checking that each generated deck contained a 60-card maindeck and a 15-card sideboard.

To calibrate new-set overlap against chance, we computed two random
baselines over the $N=261$ non-land new-set card pool. The first is the
per-pair expected overlap, $\mathbb{E}[\text{overlap}] = k K / N$,
where $k$ is the number of new-set cards in the generated deck and
$K$ is the number in the PT deck. This is the hypergeometric mean
and tells us how many new-set cards a single generated-vs-PT deck
comparison would share by chance alone, given the two decks' new-set
card counts. The second is the aggregate expectation across the
$d = 9$ decks generated for each model under each pipeline configuration,
$\mathbb{E}_{\text{agg}} = T \cdot \big[1 - \prod_{i=1}^{d}(1 - k_i/N)\big]$,
where $k_i$ is the new-set card count of the $i$-th generated deck and
$T = 19$ is the total number of distinct new-set cards in the PT
ground-truth set. This is the expected number of distinct PT new-set cards
rediscovered across the configuration's nine decks if each deck were an
independent uniform random draw from the pool, computed analytically
and via $10{,}000$ Monte-Carlo resamples for percentile bounds. It
serves as an upper bound on expected random rediscovery because the
actual generated decks within a configuration share agent~1 and agent~2
intermediates and are therefore  correlated.

\subsection{Statistical tests}

For F1, we tested the association between the quality dimensions and human-reviewed real-innovation matches. For each criterion, we collapsed the verdict to a binary pass/non-pass split (e.g.\verb| PASS+GREY/PARTIAL| vs \verb|FAIL|) and ran Fisher's exact test on the resulting $2\times 2$ contingency table against matched versus unmatched status, pooled across all six models (n = 881 ideas). Odds ratios (OR) are reported with 95\% confidence intervals (CI) derived from the standard error of the log OR. We also report cross-run idea convergence for each model. For each idea in a general run $i$, we compute its maximum cosine similarity to any idea in another general run $j \neq i$. For MTG, one-shot and tool-use pipeline configurations were compared across the six matched models using a paired Wilcoxon signed-rank test on new-set recall, and a Friedman test was used as the corresponding aggregate test on main-deck legality across all six models. For card-level analysis, we tested the rank correspondence between PT deck breadth and the number of generated decks playing each new-set card using Spearman's $\rho$ and Kendall's $\tau$ across the 19 new-set cards present in any of the 19 PT reference decks. Because generated outputs in both domains are clustered by model, run, and pipeline configuration, these tests are interpreted as exploratory associations rather than model-level causal comparisons.

\section{Results}\label{sec:results}

\subsection{F1}

\begin{figure}[t]
\centering
\includegraphics[width=\columnwidth]{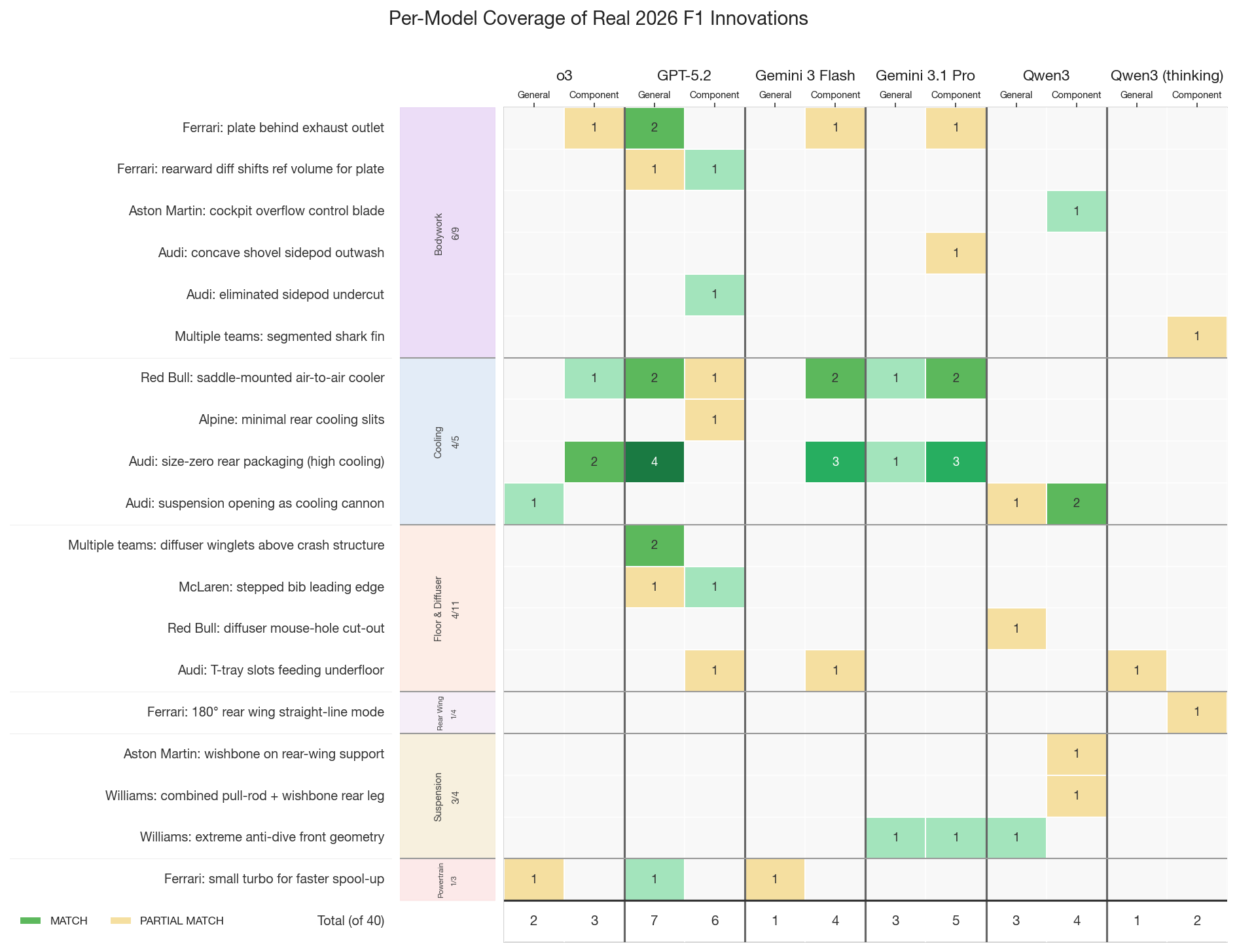}
\caption{Human-confirmed matches between model-generated ideas and real 2026 F1 innovations. Rows show the real innovations, grouped by car area, and columns show each model under general (full-car) and component-focused configurations. Green cells denote matches, yellow cells denote partial matches, with cell values indicating the number of generated ideas that matched.}
\label{fig:f1-blog-heatmap}
\end{figure}

The six models proposed 881 candidate technical innovations across both general (full-car) and component-focused runs. In total, 19 of 40 (48\%) real 2026 pre-season innovations were independently suggested by at least one model. Performance varied substantially by model, with Qwen3 235B thinking partially matching 3 real-world innovations, while GPT-5.2  matched 10. Coverage was uneven across car areas with higher recall in regions such as cooling (4/5) and bodywork (6/9). This may reflect a combination of the experimental setup, as well as text distribution in the regulations and model training data. The two pipeline configurations were complementary, with the union of their matches exceeding either method alone for all models other than Gemini 3.1 Pro. Figure~\ref{fig:f1-blog-heatmap} presents all matches between model-generated ideas and real 2026 F1 innovations, grouped by car area and separated by pipeline configuration.

\begin{table}[t]
\centering
\small
\caption{Quality signal pass rates by model. \emph{Ideas} denotes the number of candidate technical innovations generated across all runs. \emph{All-pass} is the fraction passing all three filters, while \emph{Eng.}, \emph{Cite}, and \emph{Comp.} report pass rates for engineering plausibility, citation accuracy, and rule compliance respectively.}
\label{tab:f1-per-model}
\setlength{\tabcolsep}{3pt}
\begin{tabular}{lrrrrr}
\toprule
Model & Ideas & All-pass & Eng. & Cite & Comp. \\
\midrule
GPT-5.2          & 166 & 83\% & 99\% & 93\% & 88\% \\
Gemini 3 Flash   & 149 & 49\% & 93\% & 71\% & 62\% \\
Gemini 3.1 Pro   & 108 & 41\% & 76\% & 73\% & 63\% \\
 Qwen3 235B       & 152 & 39\% & 85\% & 61\% &62\% \\
o3               & 151 & 28\% & 84\% & 50\% & 51\% \\
Qwen3 235B (think.) & 155 & 12\% & 54\% & 55\% & 41\% \\
\bottomrule
\end{tabular}
\end{table}

All three quality signals were significantly associated with human-confirmed matches. Rule compliance carried the strongest signal, with legally compliant or grey-zone ideas (n = 542) having $9\times$ higher odds of matching a real innovation than non-compliant ideas (n = 339; OR = 8.92, 95\% CI [2.74, 29.08], $p = 2.9 \times 10^{-6}$). Citation accuracy (OR = 3.11, 95\% CI [1.30, 7.46], $p = 0.007$) and engineering plausibility (OR = 4.65, 95\% CI [1.11, 19.44], $p = 0.022$) showed weaker but still significant associations. Idea counts and quality signal pass rates per-model are summarized in Table~\ref{tab:f1-per-model}.

Outputs differed in stability across independent runs. GPT-5.2 produced the most consistent output (cross-run best-match cosine similarity $\mu = 0.70$), repeatedly returning to a narrower region of the design space, while o3 explored more widely ($\mu = 0.62$). Match counts were also noisy across independent runs. The mean within-model standard deviation was 0.78, and five of six models had at least one run with zero matches. Hence, we reported the union across independent runs for each model, rather than treating any single run as representative. To check that coverage is not merely a function of generation volume, we also report distinct innovations matched per 100 ideas (\autoref{tab:f1-volume}); GPT-5.2 again ranks highest under this normalization (6.0 matches per 100 ideas), and idea count is not significantly associated with distinct matches across the six models (Spearman $\rho = 0.33$, $p = 0.52$, $n = 6$).

Qualitatively, some of the most novel ideas identified real engineering trade-offs opened up by the 2026 regulations. For example, GPT-5.2 suggested that the removal of the MGU-H could make a smaller, lighter turbocharger attractive, sacrificing some top-end performance for faster spool and improved drivability. This aligns with reports that Ferrari has pursued a similar 2026 power-unit strategy, potentially contributing to its strong race starts so far this season. In another idea, GPT-5.2 also proposed moving the gearbox differential rearward to create downstream aerodynamic freedom. Ferrari appears to have exploited a similar reference volume opportunity to place a vane behind the exhaust and improve diffuser performance. Despite not identifying all the real-world details, this illustrates that frontier models can sometimes conduct novel reasoning in complex domains. 

By contrast, weaker models more frequently generated unsafe, physically impossible, or legally implausible ideas. For example, Qwen3 proposed placing batteries within side-impact structures. However, some ideas still identified novel gaps in the regulations, such as exploiting definitions around rotating rear-wing elements, albeit with limited overlap to Ferrari's real novel 180-degree rotating rear-wing concept.

\subsection{MTG}

\begin{figure}[t]
\centering
\includegraphics[width=\columnwidth]{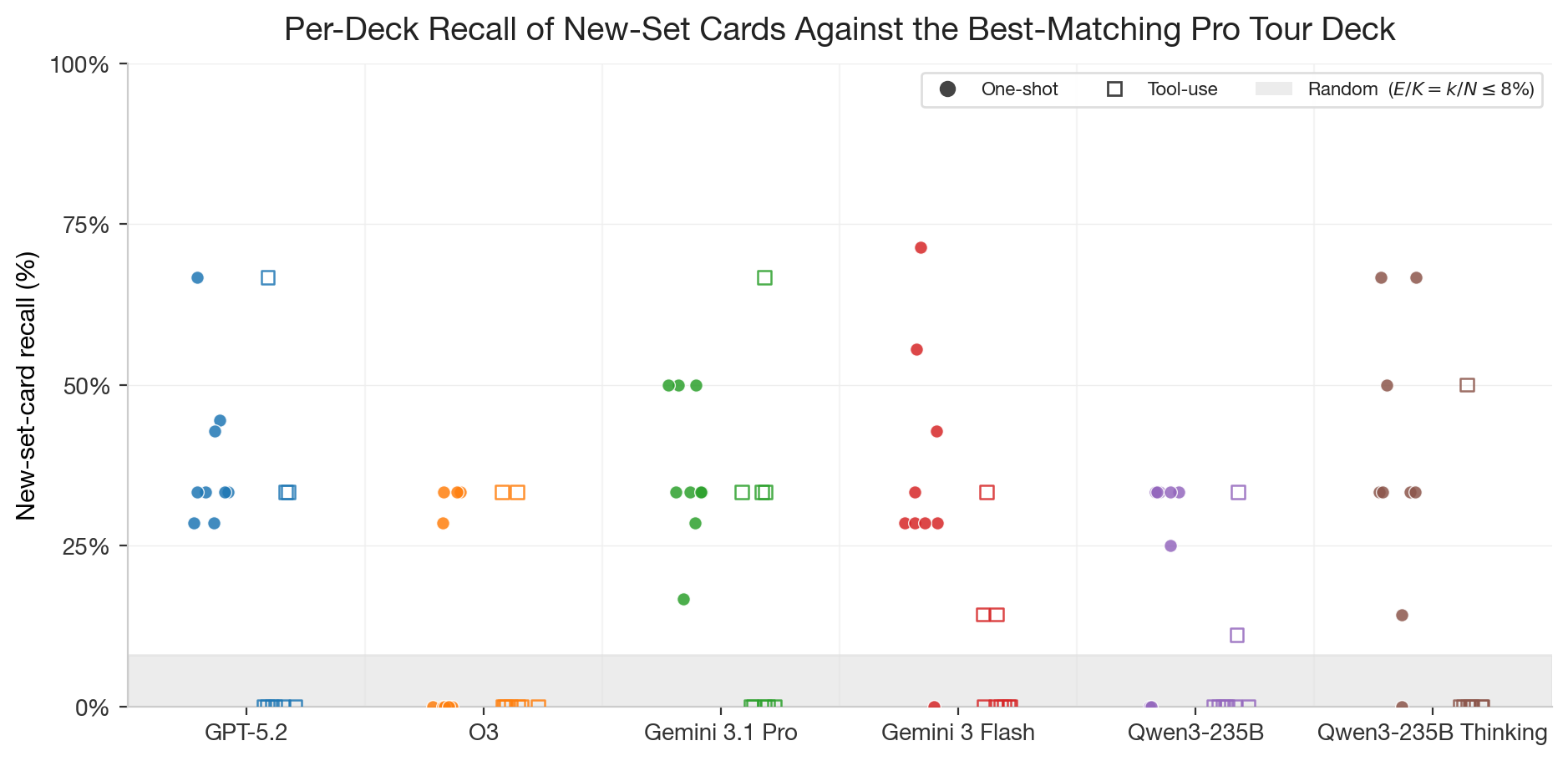}
\caption{Fraction of new cards from each generated deck's closest PT match that were correctly predicted, broken down by model and pipeline configuration. Each marker is one generated deck. The grey band marks the per-pair random expectation; points above it indicate above-chance recovery.}
\label{fig:oneshot-vs-tools}
\end{figure}

Across the six models and 108 generated decks, 14 of 19 (74\%) PT new-set cards were selected. The strongest single deck-pair match was from Gemini 3 Flash in the one-shot configuration, recovering 5 of 7 new-set cards (71\%, \autoref{fig:oneshot-vs-tools}) in the semi-finalist Izzet Elementals deck against a per-pair expectation of 0.32 cards (15.5\(\times\)). The strongest configuration across multiple decks was also one-shot Gemini 3 Flash, the only model whose nine-deck total exceeded the upper quartile of its random-baseline band in both configurations (8 of 19 one-shot, 1.34\(\times\) expected; 3 of 19 tool-use, 2.13\(\times\) expected; \autoref{fig:model-vs-random}).

At the per-deck level, 60 of 108 generated decks (56\%) achieved a best-match new-set-card recall above the per-pair random expectation (\autoref{fig:oneshot-vs-tools}). Tool-use significantly reduced new-set discovery for every model (paired Wilcoxon \(p = 0.031\); 45/54 [83\%] one-shot decks beat random versus 15/54 [28\%] for tool-use decks). The same scaffolding substantially increased main deck validity for GPT-5.2 (11\% to 100\%) and Qwen3 (0\% to 89\%), but no statistical differences were observed across all models (Friedman p = 0.24).

\begin{figure}[t]
\centering
\includegraphics[width=\columnwidth]{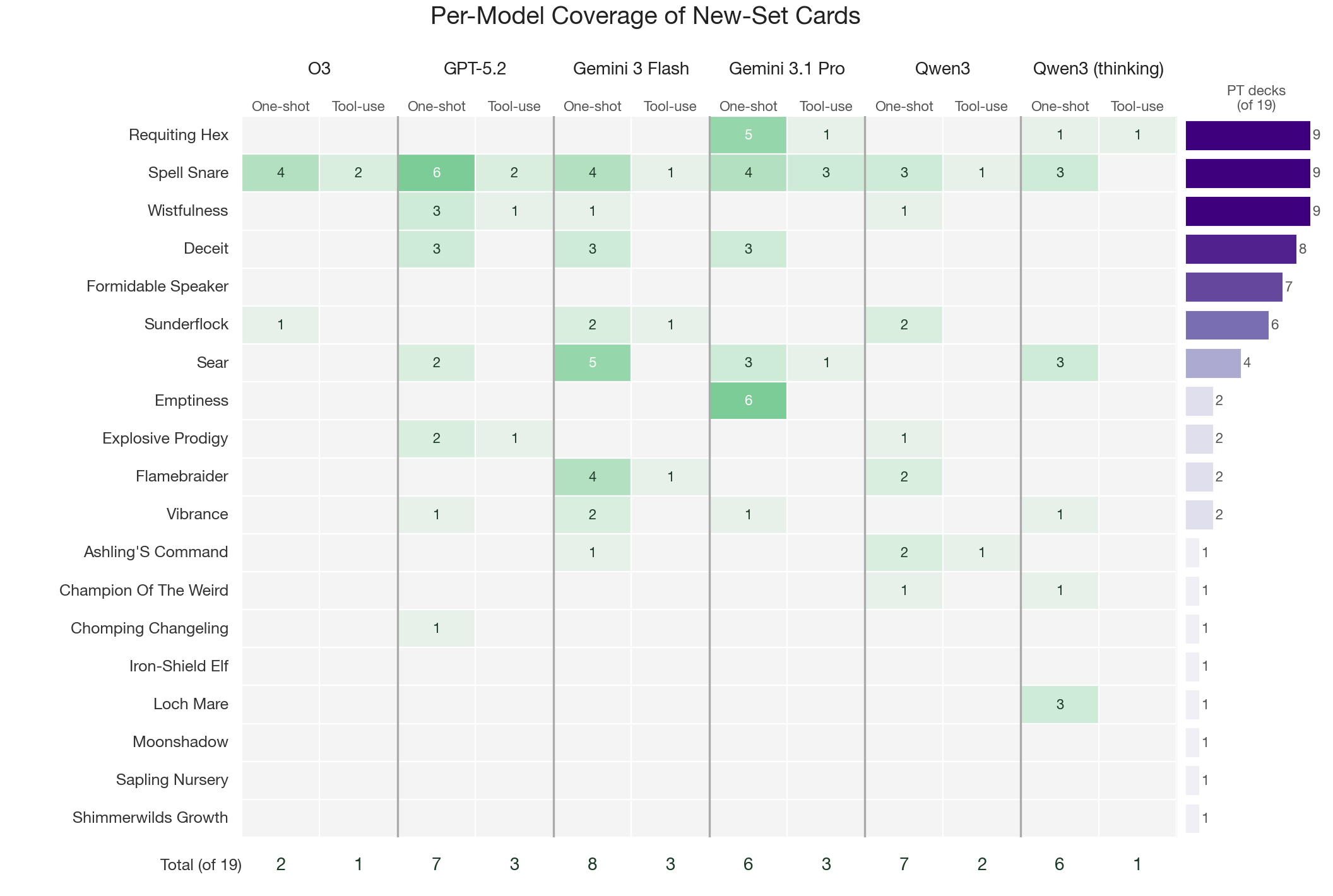}
\caption{Per-model coverage of the 19 new cards present in any of the PT decks. Rows are PT new cards, ordered by the number of ground-truth decks containing them (right bar). Both one-shot and tool-use configurations are shown for each model. Cell values give the number of generated decks (out of 9) that included the card.}
\label{fig:cardmatrix}
\end{figure}

At the card level, model selections aligned with PT adoption. Across the 19 new-set PT cards, the number of generated decks playing a card correlated strongly with the number of PT decks containing it (Spearman $\rho = 0.74$, $p = 0.0003$, $n = 19$; Kendall $\tau = 0.62$, $p = 0.0008$). Models converge on PT staples in proportion to how widely those staples are actually played, even though aggregate new-set discovery only modestly exceeded the random baseline. The exceptions to this trend are informative. Card Spell Snare appears in 33 of 108 generated decks but only 9 of 19 PT decks, and Sear in 14 generated decks against 4 PT decks, indicating that models over-predict generically powerful effects relative to their actual meta footprint. Conversely, Formidable Speaker appears in 7 of 19 PT decks but in zero generated decks, illustrating that build-around cards whose value depends on supporting deck construction are systematically missed (\autoref{fig:cardmatrix}).

\section{Discussion and Limitations}

This study argues that adversarial, fast-moving real-world domains can provide a practical intermediate test bed for evaluating AI scientist capabilities. Across both domains, frontier systems produced many plausible outputs and in several cases recovered post-cutoff expert choices. However, a difficulty appears to be discrimination, identifying ideas which are grounded, useful, and worth pursuing. The 881 candidate F1 ideas covered 19 of 40 real innovations between them, but most ideas did not match any innovation, and 89\% of pairs that the LLM novelty judge flagged for human review were downgraded. In MTG, 14 of the 19 PT new-set cards were selected by some generated deck, yet aggregate discovery within any single configuration only modestly exceeded its random baseline. This highlights that open-ended ideation benchmarks should not just reward fluency and volume, but should explicitly measure filtering, prioritization, and meta-alignment with experts. On an aggregate level there was some evidence that models converge on real-world expert choices in proportion to how widely those choices are adopted, suggesting that the gap for AI scientists lies in over-emission and selective omission rather than in absent understanding.

Models failed in the same way across both domains. They created candidates whose value is immediately obvious more readily than candidates whose value emerges only when composed with surrounding context. In MTG, generic-effect cards were over-predicted (Spell Snare appeared in 33 of 108 generated decks against 9 of 19 PT decks; Sear in 14 against 4), while build-around cards whose value emerges only from a supporting deck shell were systematically missed (Formidable Speaker in 0 generated decks against 7 of 19 PT decks). The F1 loophole taxonomy shows the analogous pattern. \texttt{INTERFACE\_COUPLING}, exploitation of cross-section regulatory junctions, was the most frequently flagged loophole type by agent~2 (378 of 1311 mappings, 28.8\%) but matched real innovations at only 5.0\%, while \texttt{BROAD\_FREEDOM}, regions where the regulation specifies what a system must do but leaves where and how unconstrained, was less frequently flagged (20.7\%) but produced the highest match rate (7.7\%). This failure pattern is highly relevant to scientific discovery, as the value of a hypothesis often depends on how it interacts with the rest of a research program and breakthroughs in science often cannot be planned \citep{stanley2015greatness}.

The two domains explored also illustrate a trade-off between scaffolding and exploration. Component-focused prompting in F1 increased idea coverage, suggesting that narrowing the search space or deliberately diversifying generation can expose new ideas. In MTG, by contrast, tool-use scaffolding improved legality for some models but reduced new-set discovery for every model (paired Wilcoxon $p = 0.031$). This is a useful insight for AI scientist systems. More structure can make outputs easier to validate and less likely to violate formal constraints, but it may also push systems toward conservative or locally consistent solutions. This, alongside the observation that the best-performing model (GPT-5.2) was also the most narrowly-clustered in the idea space (cross-run convergence $\mu = 0.70$), may help explain why in this work we did not observe a consistent differentiation between reasoning and non-reasoning models. However, only a small set of models was used in this research and so this requires further investigation.

This work has several limitations. First, a match shows convergence with a human idea under shared constraints. It is not evidence that the model reasoned the same way the expert did, nor that the idea would translate to provide genuine value. Second, expert artifacts are an incomplete ground truth. Public F1 coverage overrepresents visible aerodynamics and packaging while many decisive sources of performance remain confidential, and MTG decklists reveal successful tournament choices but do not capture all viable decks, testing processes, or private strategic reasoning. Third, this framework can test novelty and reasoning under constraints, but the domains evaluated are still proxies for open-ended scientific discovery. Finally, although this framework can be run prospectively, the human artifacts in this study had already been generated at the time of evaluation. Future instantiations should pre-register prompts, corpus, models, and runs before expert artifacts are released.

The framework presented in this work is portable. Any adversarial fast-moving domain with publicly observable expert outputs and a clear information cutoff can be instantiated as a benchmark. In practice, the most tractable domains would have periodic, well-defined reveal events that frequently re-instantiate the benchmark at low marginal cost, a calibratable random baseline against which matches can be judged, and feature many independent experts whose breadth of ideas can give a signal of which matter most. Other candidate domains include legal and financial regulatory cycles. How to aggregate metrics from heterogeneous domains into a single interpretable score remains an open question. The most useful AI scientist benchmarks may combine prospective public time-delayed artifacts with richer process data: what information was available at the cutoff, what candidate ideas were considered, why some were discarded, and what evidence later confirmed or falsified them. In scientific practice, this might correspond to pre-registering experimental designs before a study runs, shortlisting candidate targets before a drug-discovery campaign, or proof-search strategies before a conjecture is formally settled. Such tasks would preserve the desirable properties used here, namely complex design spaces, independent expert search, and delayed verifiable outcomes, while moving closer to real scientific contribution. Future instantiations may also benefit from coupling AI scientist systems to simulation, CFD, or laboratory automation, since several F1 failure modes appear to reflect the absence of physical-validation tools \citep{pittphilsci28024}.

Time-delayed real-world evaluation allows for reference artifacts to be produced independently by human experts after the model knowledge cutoffs. The results in this paper support a cautious but constructive view of current AI scientist capabilities. Frontier models can sometimes reason into the same regions of a complex design space as expert humans. However, their outputs remain noisy, weakly calibrated, and dependent on evaluation choices. Progress will require systems that not only generate hypotheses, but also attach trustworthy provenance, check constraints, estimate feasibility, seek evidence, and decide which ideas deserve scarce experimental attention.

\section{Conclusion}

Benchmarking AI scientists and verifying their discoveries will likely become the bottleneck to safely deploying autonomous research systems and achieving accelerated scientific progress. Complex, adversarial real-world domains that are publicly accessible and fast-moving could act as a practical tool to evaluate such systems and assess their capacity to be innovative before deploying systems into higher-stakes scientific settings. 

\section*{Code and data availability}

Code, data, prompts, and generated outputs are available for the F1 domain at \url{https://github.com/WilliamBolton/f1-ai-scientist-paper} and for the MTG domain at \url{https://github.com/WilliamBolton/mtg-AI-scientist-paper}. The full pipeline is reproducible from these repositories; the only stochasticity in the experiments comes from the sampler inside each LLM call.

\section*{Impact Statement}

This paper proposes a framework for evaluating the capabilities of AI scientists. Better evaluation could help researchers identify which AI scientist systems are genuinely useful, where they fail, and when they independently produce useful contributions. Potential risks include over-claiming model capabilities, encouraging benchmark gaming, or transferring conclusions from proxy domains too directly to scientific practice. We therefore emphasize that this evaluation framework should complement, not replace, expert judgment, and external validation. The domains studied here are public and do not involve human subjects or sensitive personal data. They remain proxies for scientific discovery and should be interpreted accordingly.
\section*{Acknowledgements}
This work was funded by ARIA, DSIT and Pillar VC under the Encode: AI for Science Fellowship.

\bibliography{paper}
\bibliographystyle{icml2026}

\appendix
\onecolumn

\section*{Supplementary Methods}

\subsection*{Models}

Six frontier LLMs spanning three providers were evaluated. The OpenAI models GPT-5.2 (August 2025 knowledge cutoff) and o3 (June 2024 knowledge cutoff) were accessed via the OpenAI API. The Google models Gemini 3 Flash and Gemini 3.1 Pro (both January 2025 knowledge cutoff) and the Alibaba models Qwen3 235B and Qwen3 235B (thinking) (both June 2025 knowledge cutoff) were accessed via OpenRouter. Every cutoff predates both the FIA 2026 Section C Technical Regulations (released 10 December 2025) and Lorwyn Eclipsed (released 23 January 2026), so neither corpus could have appeared in any model's training data. F1 generated ideas and real innovations were embedded with OpenAI's \texttt{text-embedding-3-small} model.

\subsection*{Ground-truth collection}

The reference set of 40 real 2026 pre-season innovations was compiled from publicly available technical analysis published between January and March 2026 (\autoref{tab:f1-sources}). Sources span written publications and technical videos covering pre-season testing and car launches. An LLM extraction pipeline converted source text and video transcripts into structured entries, each carrying \texttt{innovation\_id}, \texttt{headline}, \texttt{description}, \texttt{technical\_mechanism}, \texttt{lay\_summary}, \texttt{car\_area}, the team or teams reported as using the innovation, a confidence flag, and source-citation fields.

\begin{table}[H]
\centering
\small
\begin{tabular}{p{0.18\textwidth}p{0.10\textwidth}rp{0.44\textwidth}}
\toprule
\textbf{Source} & \textbf{Date} & \textbf{Innovations} & \textbf{URL} \\
\midrule
YouTube & 2026-02-23 & 9 & \href{https://www.youtube.com/watch?v=hs7PcFX-Tmw}{youtube.com/watch?v=hs7PcFX-Tmw} \\
PlanetF1 & 2026-02-16 & 8 & \href{https://www.planetf1.com/features/f1-2026-bahrain-testing-analysis-red-bull-ferrari-mercedes-technical}{Bahrain testing analysis} \\
YouTube & 2026-02-23 & 7 & \href{https://www.youtube.com/watch?v=x7HE73ERnAE}{youtube.com/watch?v=x7HE73ERnAE} \\
YouTube & 2026-02-23 & 5 & \href{https://www.youtube.com/watch?v=uwjSzkFSumc}{youtube.com/watch?v=uwjSzkFSumc} \\
YouTube & 2026-02-23 & 4 & \href{https://www.youtube.com/watch?v=E8j2jc4ezus}{youtube.com/watch?v=E8j2jc4ezus} \\
YouTube & 2026-02-23 & 4 & \href{https://www.youtube.com/watch?v=Lf9ZIVMoPIw}{youtube.com/watch?v=Lf9ZIVMoPIw} \\
Autosport & 2026-01-31 & 1 & \href{https://www.autosport.com/f1/news/newey-extreme-aston-red-bull-mercedes-early-technical-trends-f1-2026/10794309/}{Newey-extreme early technical trends} \\
Autosport & 2026-02-23 & 1 & \href{https://www.autosport.com/f1/news/controversial-2026-f1-engine-loophole-wont-be-resolved-before-australian-gp/10792200}{2026 engine loophole article} \\
ScuderiaFans & 2026-01-28 & 1 & \href{https://scuderiafans.com/2026-f1-technical-analysis-the-most-interesting-aerodynamic-ideas-from-barcelona-testing/}{Barcelona testing aerodynamic ideas} \\
\midrule
\textbf{Total} & & \textbf{40} & \\
\bottomrule
\end{tabular}
\caption{Provenance of the 40 real 2026 pre-season F1 innovations used as the ground-truth reference set. Sources are public technical analyses published between 28 January 2026 and 23 February 2026, comprising five YouTube videos (contributing $29$ innovations) and four written articles (contributing $11$ innovations) for a total of 40.}
\label{tab:f1-sources}
\end{table}

The 19 PT decks comprise the Top 15 finishers from the Day 2 standings of PT Lorwyn Eclipsed, plus four featured decks selected for innovative approaches (\autoref{tab:mtg-sources}). Each deck was canonicalized against Scryfall to a list of (card name, count) pairs partitioned into a 60-card maindeck and a 15-card sideboard, alongside summary fields that mirror the generated-deck schema (\texttt{lay\_summary}, \texttt{game\_plan}, \texttt{key\_cards}, \texttt{key\_synergies}, \texttt{metagame\_role}, \texttt{strategy\_archetype}). Card names were normalized to Scryfall's canonical lowercase form so that overlap counts between generated and ground-truth decks are insensitive to capitalization or printing variants.

\begin{table}[H]
\centering
\small
\begin{tabular}{p{0.30\textwidth}rp{0.30\textwidth}}
\toprule
\textbf{Source} & \textbf{Decks} & \textbf{URL / coverage} \\
\midrule
PT Lorwyn Eclipsed (Day 2 Top 15) & 15 & \href{https://magic.gg/events/pro-tour-lorwyn-eclipsed}{magic.gg/events/pro-tour-lorwyn-eclipsed} \\
PT Lorwyn Eclipsed (featured archetype-diverse builds) & 4 & \href{https://magic.gg/events/pro-tour-lorwyn-eclipsed}{magic.gg/events/pro-tour-lorwyn-eclipsed} \\
\midrule
\textbf{Total} & \textbf{19} & \\
\bottomrule
\end{tabular}
\caption{Provenance of the 19 PT Lorwyn Eclipsed reference decks. The Top 15 by Day 2 standings and four featured archetype-diverse builds were collected from official PT Lorwyn Eclipsed coverage (30 January -- 1 February 2026). Each deck was canonicalized against Scryfall to a 60-card maindeck plus 15-card sideboard with all card names normalized to Scryfall's canonical lowercase form.}
\label{tab:mtg-sources}
\end{table}

\subsection*{Run structure}

For F1, each model produced three independent general runs and one run per car area in the by-component configuration. General runs use the same prompt across all three repetitions, while by-component runs use a per-area prompt that restricts agent~2's scope to the target car area's articles and adjacent interfaces. For MTG, each run of the three-agent pipeline produces three decks under different strategic angles, and each model under each pipeline configuration was run three times, yielding 18 generated decks. Tool-use runs share the agent~1 and agent~2 stages with one-shot but agent~3 builds each deck through a tool-calling loop with three primitives (\texttt{add\_card}, \texttt{remove\_card}, \texttt{finish\_deck}).

\subsection*{F1 loophole tagging and quality verdict scales}
\label{loophole-tagging}

Agent~2 emits a structured output containing 15 to 20 degrees of freedom, each carrying a \texttt{loophole\_type} drawn from \{\texttt{INTERFACE\_COUPLING}, \texttt{DEFINITION\_EDGE\_CASE}, \texttt{BROAD\_FREEDOM}, \texttt{NOT\_PROHIBITED}, \texttt{EXCEPTION\_ZONE}\} and a list of cited article numbers. Each generated idea inherits the union of loophole types from those mappings whose article identifiers overlap with the idea's \texttt{enabling\_articles}. Tags are non-exclusive, and an idea is tagged with on average 1.94 categories (range 1 to 4).

The three quality judges score each generated idea on independent rubrics. Citation accuracy is one of \texttt{PASS} (every cited article exists and substantively supports the claim), \texttt{PARTIAL} (mostly correct, minor omissions or slips), or \texttt{FAIL}. Rule compliance is one of \texttt{PASS} (clearly legal under the regulations), \texttt{GREY} (regulation is silent or admits multiple interpretations), or \texttt{FAIL} (violates a specific article). Engineering plausibility is binary, \texttt{PASS} or \texttt{FAIL}, based on adhering to basic physics. The all-pass aggregate reported in Table 1 admits an idea if its citation verdict is \texttt{PASS} or \texttt{PARTIAL}, its compliance verdict is \texttt{PASS} or \texttt{GREY}, and its engineering verdict is \texttt{PASS}.

\subsection*{Human review protocol}

All 519 candidate pairs that survived the shortlist filter were reviewed by a single author acting as the adjudicator. The reviewer is not a domain expert in either Formula 1 engineering or competitive Magic. For each candidate the reviewer saw the generated idea (\texttt{concept\_description}, \texttt{performance\_mechanism}, \texttt{lay\_summary}, \texttt{regulatory\_hook}), the candidate real innovation (\texttt{headline}, \texttt{technical\_mechanism}, \texttt{lay\_summary}, \texttt{car\_area}, \texttt{teams}), and the LLM judge's free-text rationale, but not its discrete verdict label, so that the reviewer's decision was not anchored on the upstream judge. A pair was assigned \texttt{MATCH} when the generated idea's \texttt{performance\_mechanism} described the same physical effect as the real innovation's \texttt{technical\_mechanism} with similar geometry, and \texttt{PARTIAL} when the generated idea described the same physical effect via different geometry, or similar geometry achieving a related but distinct effect. Sharing only the same regulatory area or the same problem framing was explicitly insufficient for \texttt{PARTIAL}. All headline counts and the four Fisher's exact tests in Section 5.1 use the post-review labels.

\section*{Supplementary Results}

\subsection*{F1}

Among 519 human-reviewed pairs, admitted into the review pool via a dual top-3 cosine shortlisting rule, cosine similarity between generated \texttt{performance\_mechanism} and real \texttt{technical\_mechanism} did not discriminate matched ($n = 55$, mean $= 0.628$) from unmatched ($n = 464$, mean $= 0.621$) pairs (Mann-Whitney one-sided $p = 0.22$), and the match rate stayed flat across similarity bins (9 to 12\% in [0.50, 0.70), and 9\% in [0.70, 0.75)). Embedding similarity therefore acts as the recall mechanism that builds the shortlist, not as a discriminative signal within it. The full per-loophole-type distribution and match rates are given in \autoref{tab:f1-loophole}.

\begin{figure}[H]
\centering
\includegraphics[width=0.6\columnwidth]{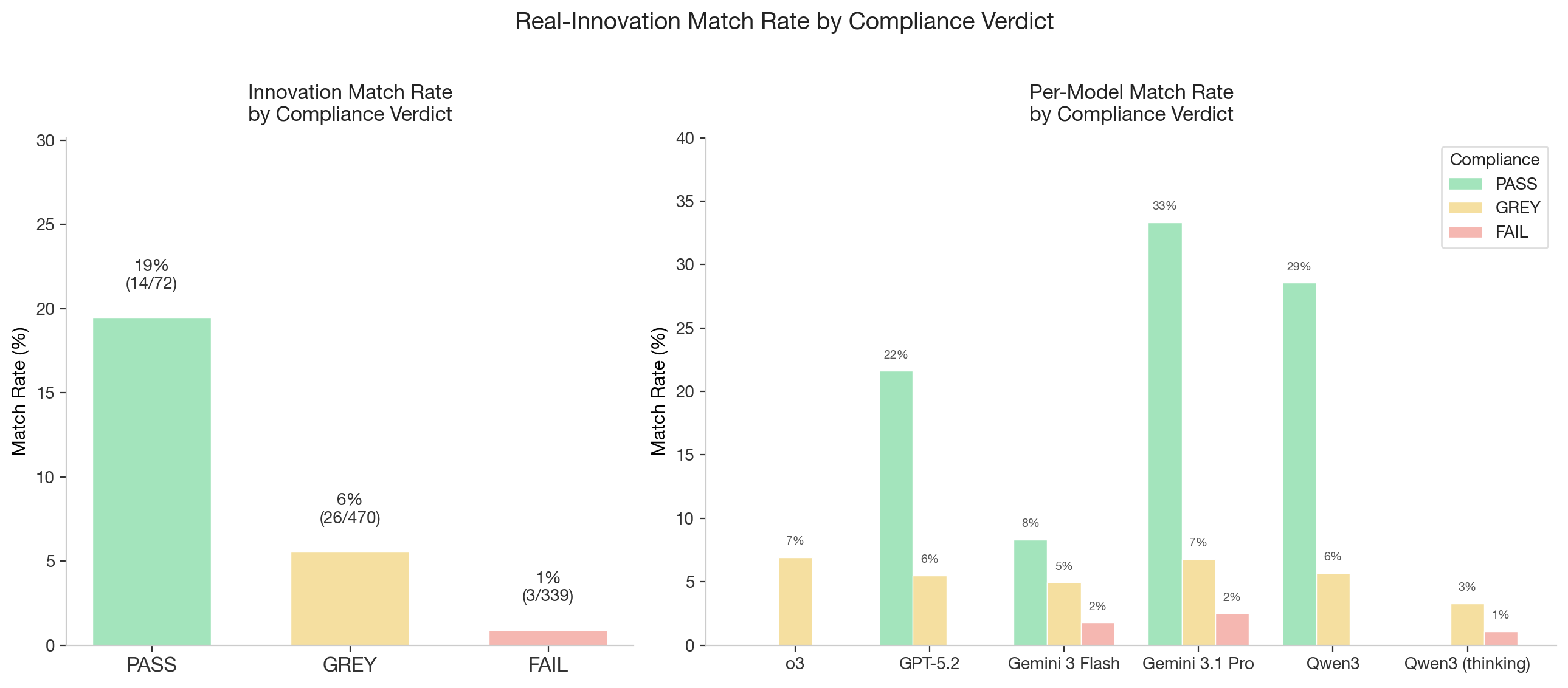}
\caption{Match rate by LLM-judge compliance verdict. The left panel shows the pooled rate across all six models. The right panel shows per-model rates broken down by compliance verdict.}
\label{fig:f1-compliance-match-rate}
\end{figure}

\begin{figure}[H]
\centering
\includegraphics[width=0.6\columnwidth]{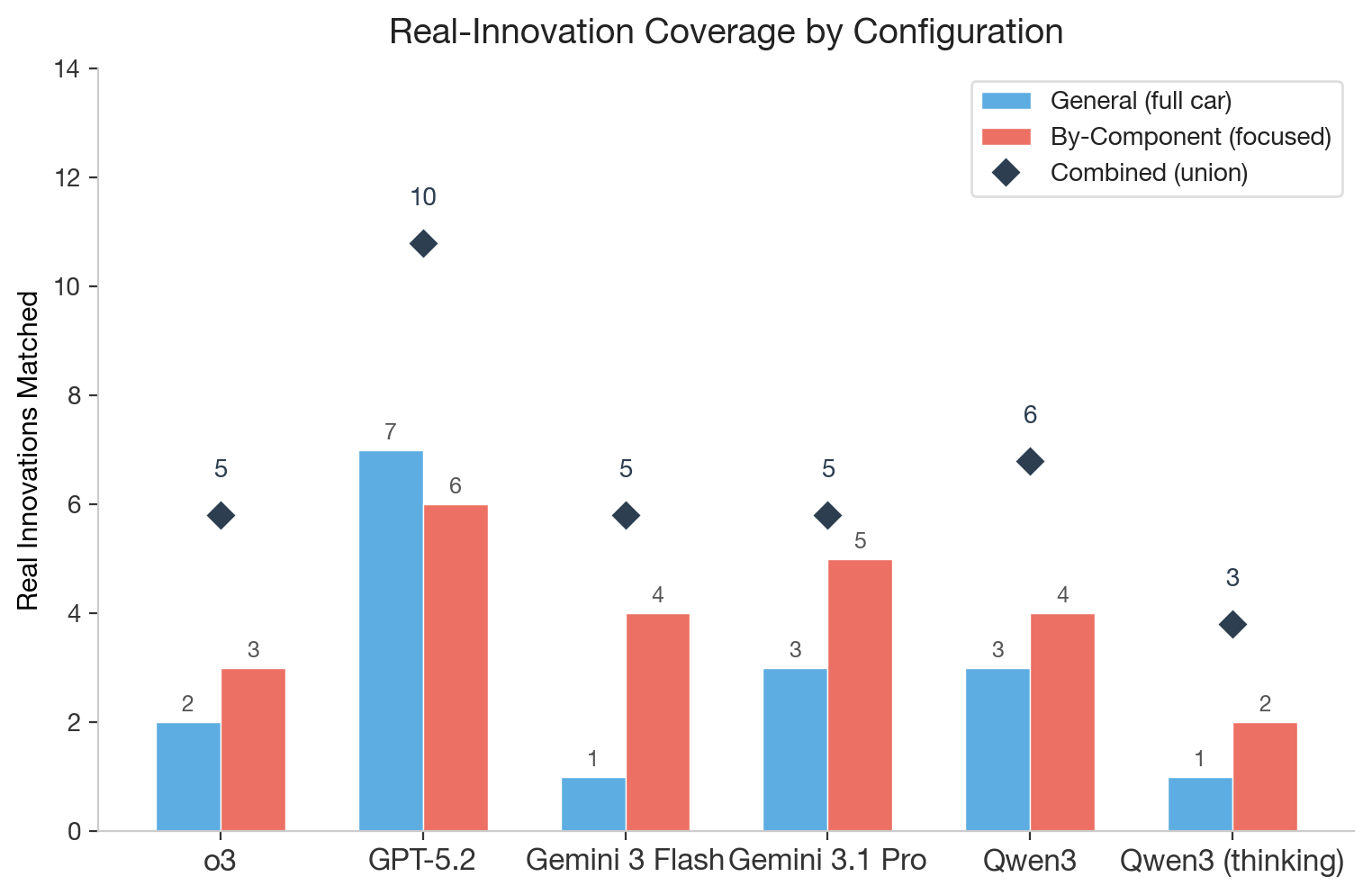}
\caption{Real-innovation coverage by configuration. For each model the blue bars show distinct real innovations matched via three independent general (full-car) prompt runs, the red bars via the component-focused prompt set (one run per car area), and the diamond marker denotes the union across both configurations.}
\label{fig:f1-config-coverage}
\end{figure}

\begin{figure}[H]
\centering
\includegraphics[width=0.6\columnwidth]{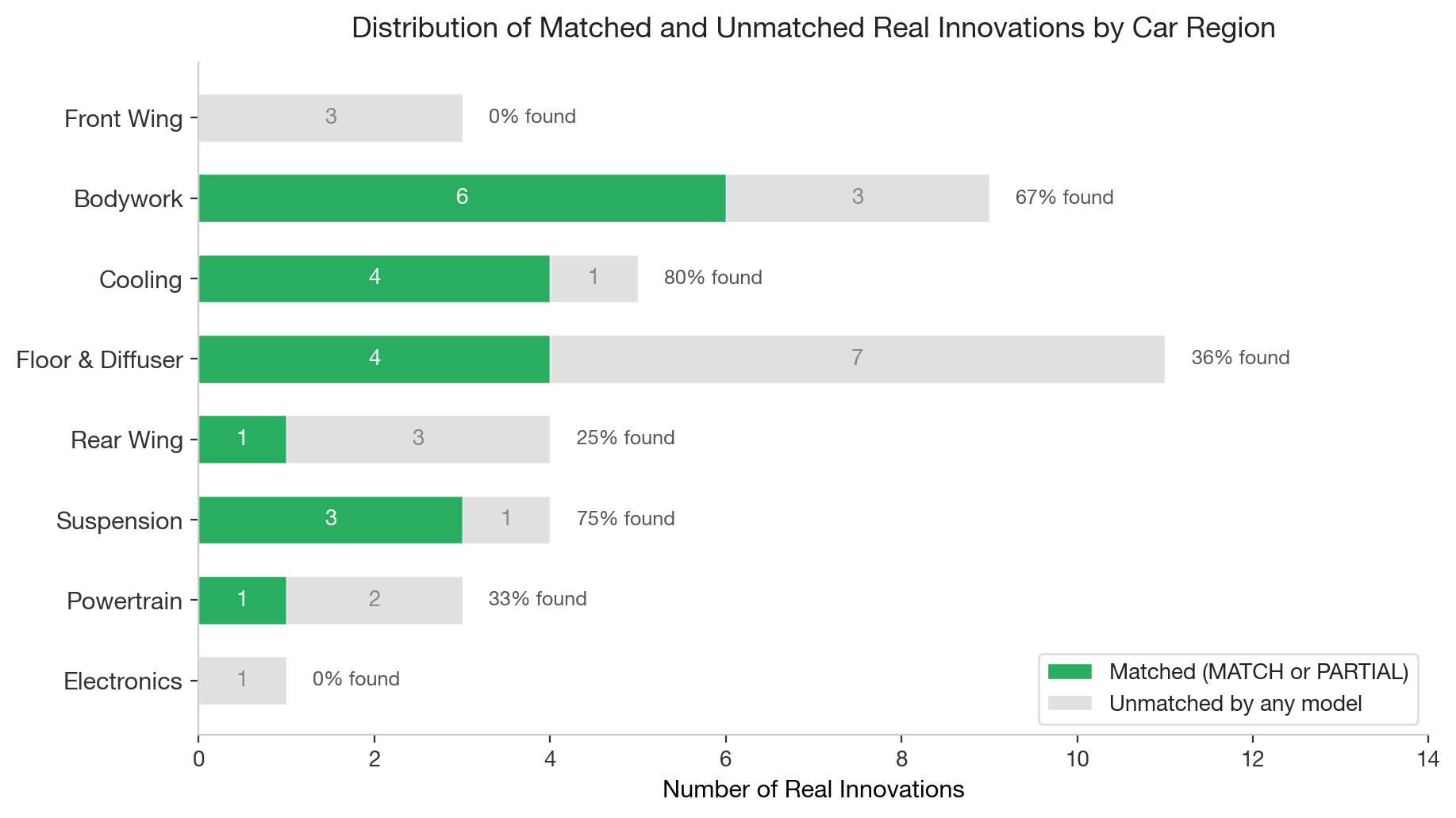}
\caption{Distribution of matched and unmatched real innovations by car region. Each row is one of the eight regions used by both the model-generated and real-innovation schemas. Green segments count innovations that any of the six models matched (\texttt{MATCH} or \texttt{PARTIAL} after human review), grey segments count innovations no model matched.}
\label{fig:f1-unmatched-area}
\end{figure}

Pooled across all six models, ideas judged \texttt{PASS} on rule compliance match real innovations at 19.4\% (14 of 72), \texttt{GREY} at 5.5\% (26 of 470), and \texttt{FAIL} at 0.9\% (3 of 339), as shown in \autoref{fig:f1-compliance-match-rate}. The pooled $2\times 2$ Fisher's exact test contrasting \texttt{PASS} or \texttt{GREY} ideas against \texttt{FAIL} ideas yields OR $= 8.92$, 95\% CI $[2.74, 29.08]$, $p = 2.9 \times 10^{-6}$. The two configurations were complementary rather than nested. Three innovations were found only via general runs and eight only via component-focused runs (\autoref{fig:f1-config-coverage}). Coverage was heavily skewed across car regions (\autoref{fig:f1-unmatched-area}): cooling (4 of 5, 80\%) and suspension (3 of 4, 75\%) were the strongest, while front wing (0 of 3) and electronics (0 of 1) were entirely uncovered. The three unmatched front-wing innovations all concern active-flap actuation packaging, a solution that no model proposed.

The evaluation pipeline itself is noisy. Of the 519 pairs the LLM novelty judge flagged for human review, 89\% were downgraded to \texttt{NO\_MATCH}, indicating that semantic similarity can confuse superficial overlap with shared mechanism and LLM judges can fail to distinguish nuanced differences.

\begin{figure}[H]
\centering
\includegraphics[width=0.6\columnwidth]{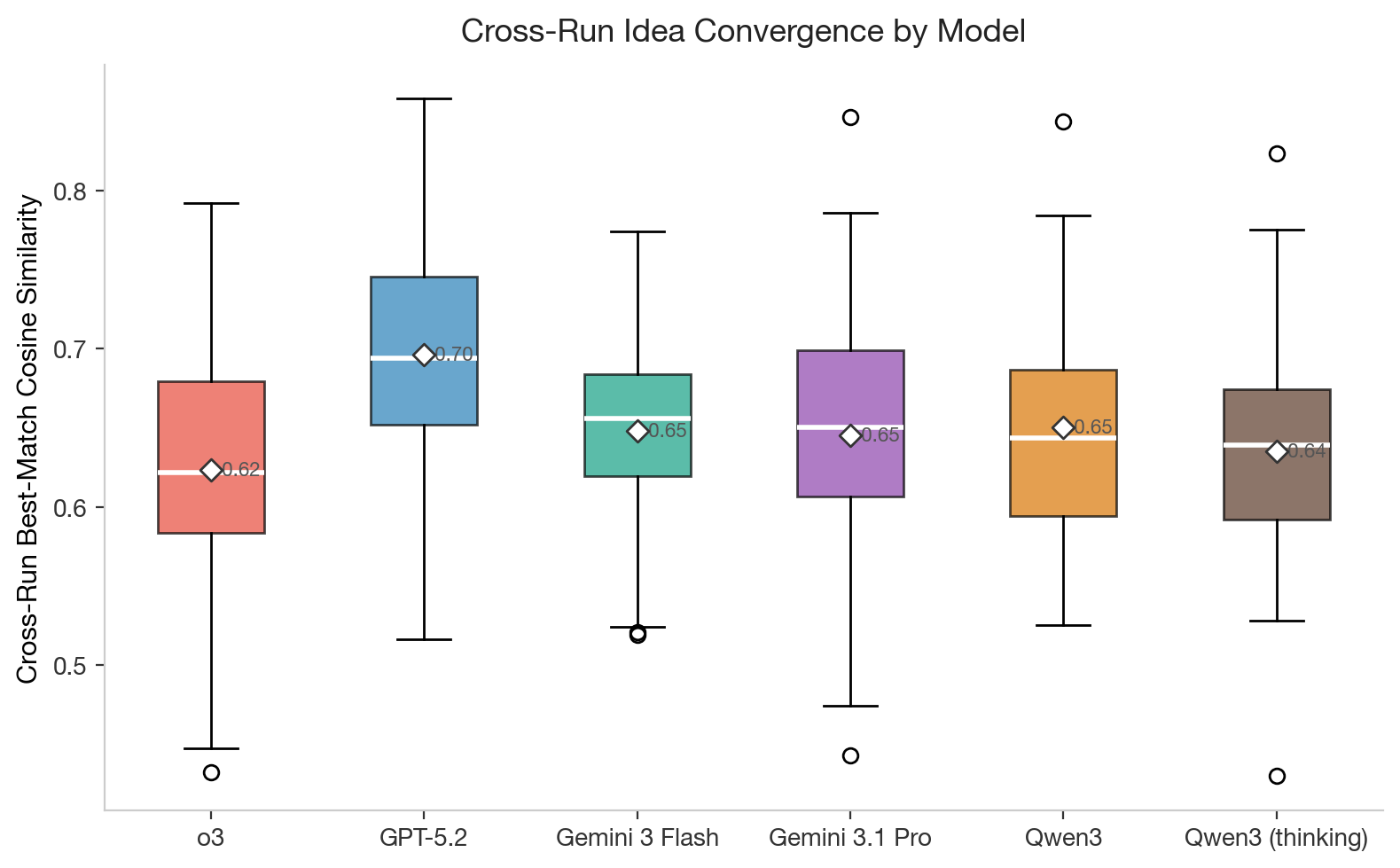}
\caption{Cross-run idea convergence by model. For each idea in run $i$ the highest cosine similarity to any idea in run $j \neq i$ is computed using \texttt{lay\_summary} embeddings. Box plots show the distribution of these best-match similarities pooled across all run pairs for each model's three independent general runs, with white diamonds denoting the means.}
\label{fig:f1-run-convergence}
\end{figure}

Union coverage rewards breadth, so we additionally normalize each model's distinct-innovation count by the number of ideas it generated (\autoref{tab:f1-volume}). GPT-5.2 ranks highest on this volume-normalized measure (6.0 matches per 100 ideas) as well as on absolute coverage, and across the six models idea count is not significantly associated with the number of distinct matches (Spearman $\rho = 0.33$, $p = 0.52$, $n = 6$). Generating more ideas does broaden coverage in absolute terms: for GPT-5.2 the union of its general and by-component runs (10 innovations) exceeds either configuration alone, with the by-component runs contributing 122 of its 166 ideas. This gain, however, likley comes from differential prompting that steers generation into distinct regions of the regulations rather than from volume alone. GPT-5.2's three general runs were the most clustered of any model (cross-run cosine $\mu = 0.70$, \autoref{fig:f1-run-convergence}), so repeatedly sampling the same prompt yields diminishing diversity. Broad coverage of ideas from models therefore depends on deliberately diversifying generation, and effectively exploring the potential ideation space, not only on increasing test time compute.

\begin{table}[H]
\centering
\small
\begin{tabular}{lrrrrrrrr}
\toprule
\textbf{Loophole type} & \textbf{o3} & \textbf{GPT-5.2} & \textbf{Gem.~Flash} & \textbf{Gem.~Pro} & \textbf{Qwen3} & \textbf{Qwen3 (T)} & \textbf{Total (\%)} & \textbf{Match rate} \\
\midrule
\texttt{INTERFACE\_COUPLING}     & 61 & 83 & 45 & 54 & 73 & 62 & 378 (28.8\%) & 5.0\%  \\
\texttt{DEFINITION\_EDGE\_CASE}  & 42 & 60 & 51 & 47 & 34 & 44 & 278 (21.2\%) & 4.2\%  \\
\texttt{BROAD\_FREEDOM}          & 40 & 36 & 47 & 41 & 61 & 46 & 271 (20.7\%) & \textbf{7.7\%} \\
\texttt{NOT\_PROHIBITED}         & 38 & 36 & 26 & 30 & 38 & 39 & 207 (15.8\%) & 5.3\%  \\
\texttt{EXCEPTION\_ZONE}         & 30 & 32 & 21 & 38 & 24 & 32 & 177 (13.5\%) & 4.6\%  \\
\midrule
\textbf{Total mappings}          & 211 & 247 & 190 & 210 & 230 & 223 & \textbf{1311} & 5.4\% \\
\bottomrule
\end{tabular}
\caption{Regulatory-loophole types identified by agent~2 across all six models. Tags are non-exclusive, and hence their sum exceeds the idea count of 881. The Match rate column gives the proportion of generated ideas that were tagged with each type via overlap with the idea's enabling article, which human review confirmed as matching a real innovation. The five categories are \texttt{INTERFACE\_COUPLING} (junctions between regulation sections, e.g.\ cooling, aero, and floor), \texttt{DEFINITION\_EDGE\_CASE} (boundary or ambiguous geometry in a definition), \texttt{BROAD\_FREEDOM} (regulations specifying what a system must do but leaving where and how unconstrained), \texttt{NOT\_PROHIBITED} (conspicuous absence of a prohibition), and \texttt{EXCEPTION\_ZONE} (explicit ``except at'' carve-outs). Match-rate differences rest on small cell counts (11 to 30 matches per type) and tags are non-exclusive, so the table is descriptive rather than inferential.}
\label{tab:f1-loophole}
\end{table}

\begin{table}[H]
\centering
\small
\caption{Generation-volume-normalized F1 coverage. \emph{Matched} is the number of distinct real innovations recovered (union across runs and configurations); \emph{Matches/100} normalizes by total ideas generated. The pooled row counts distinct innovations matched by at least one model (19 of 40); per-model values are each model's own union and overlap across models, so they do not sum to the pooled total. GPT-5.2 leads on both raw and volume-normalized coverage.}
\label{tab:f1-volume}
\begin{tabular}{lrrr}
\toprule
Model & Ideas & Matched & Matches/100 \\
\midrule
GPT-5.2             & 166 & 10 & \textbf{6.0} \\
Gemini 3.1 Pro      & 108 & 5  & 4.6 \\
Qwen3 235B          & 152 & 6  & 3.9 \\
Gemini 3 Flash      & 149 & 5  & 3.4 \\
o3                  & 151 & 5  & 3.3 \\
Qwen3 235B (think.) & 155 & 3  & 1.9 \\
\midrule
Pooled              & 881 & 19 & 2.2 \\
\bottomrule
\end{tabular}
\end{table}

\subsection*{MTG}

\begin{figure}[H]
\centering
\includegraphics[width=0.6\columnwidth]{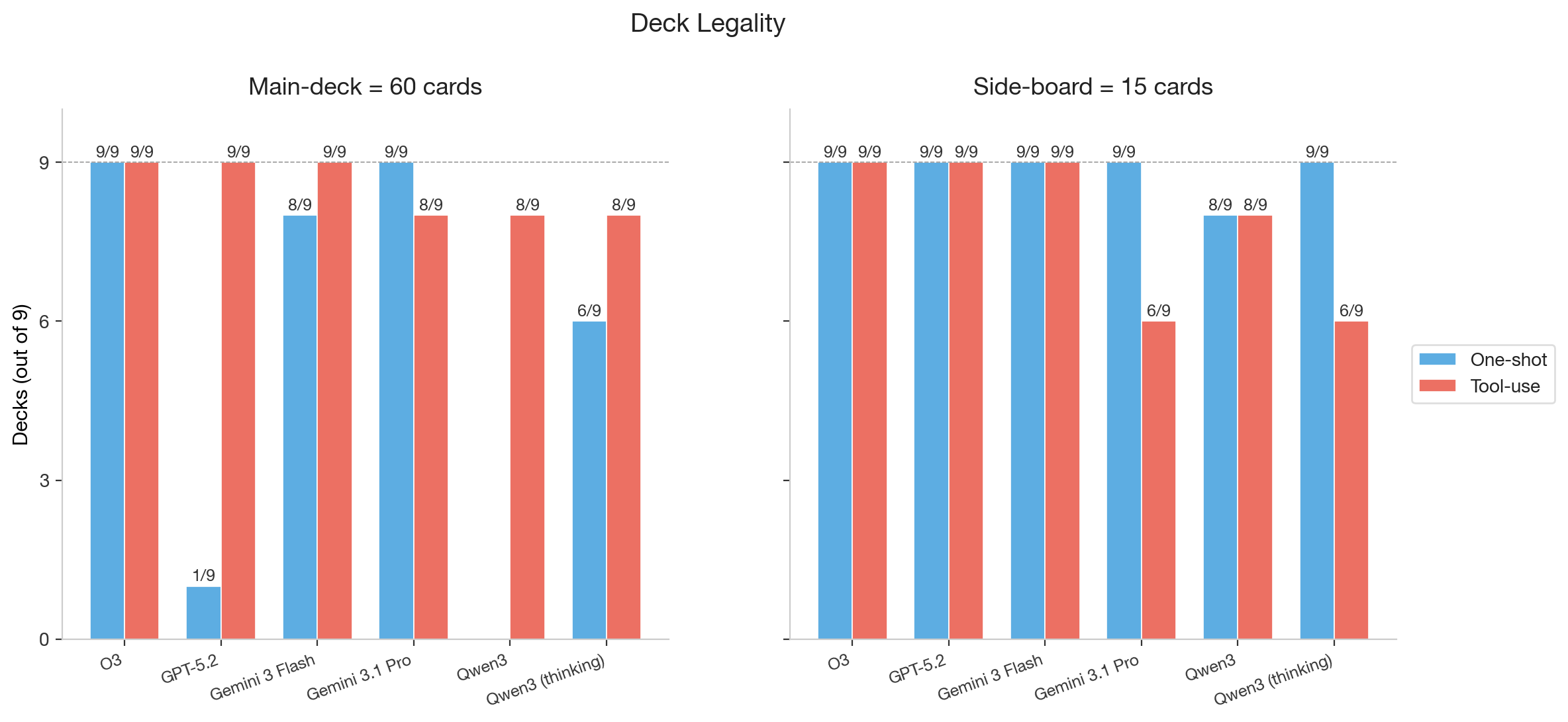}
\caption{Deterministic legality of generated decks by model and pipeline configuration. Left: decks (out of 9) with a valid 60-card main deck. Right: decks with a valid 15-card sideboard. Tool-use scaffolding substantially increases main-deck validity for GPT-5.2 (11\% $\rightarrow$ 100\%) and Qwen3 (0\% $\rightarrow$ 89\%), but no statistical difference is observed across all six models (Friedman $p = 0.24$).}
\label{app:deck-validity}
\end{figure}

Tool-use scaffolding affects main-deck legality very differently across models. GPT-5.2 jumps from 11\% to 100\% under tool-use and Qwen3 from 0\% to 89\%, while o3 and the Gemini models are already near the ceiling under the one-shot configuration (\autoref{app:deck-validity}). Sideboard legality is uniformly high across all configurations, suggesting the 15-card constraint is easier to satisfy than the 60-card main-deck constraint which is not surprising. The aggregate Friedman test across all six models is not significant ($p = 0.24$), reflecting heterogeneous sensitivity to scaffolding.

\begin{figure}[H]
\centering
\includegraphics[width=0.6\columnwidth]{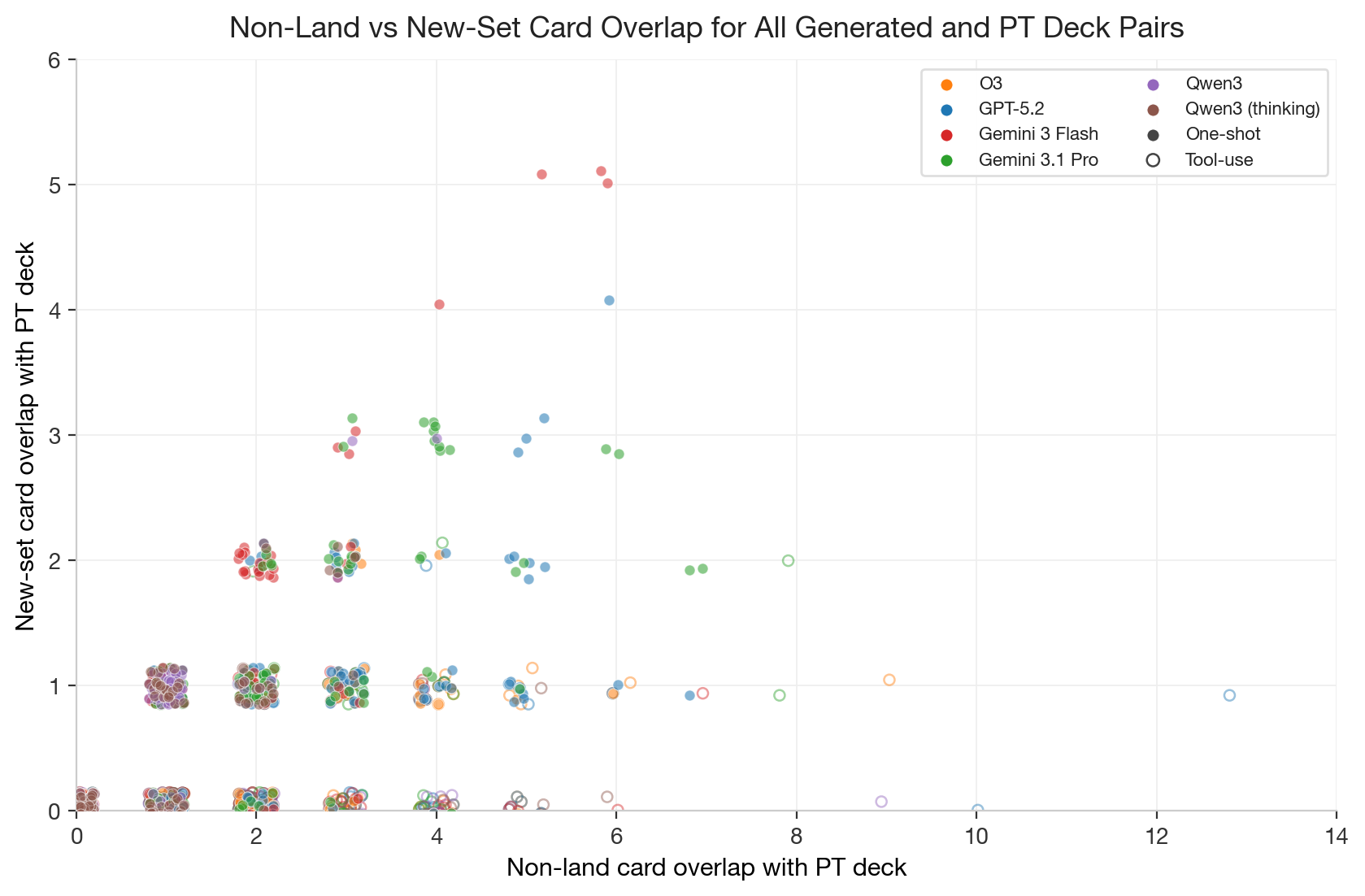}
\caption{Every generated PT deck pair, with non-land card overlap on the $x$-axis and new-card overlap on the $y$-axis. Filled markers are one-shot generations, hollow markers are tool-use, colored by model. Points are jittered for visibility.}
\label{app:best-pair-scatter}
\end{figure}

Pair-level overlap between generated decks and PT decks is concentrated in the low-overlap regime. Most of the 2{,}052 generated $\times$ PT comparisons share 0 to 3 non-land cards and 0 to 1 new-set cards (\autoref{app:best-pair-scatter}). The upper right region of the plot is empty, with no pair achieving both broad non-land alignment and high new-set recovery against the same PT deck. This suggests that models which approximate a deck's overall shape tend not to recover its specific new-set cards. One-shot pairs (filled markers) also extend further along the non-land axis than tool-use pairs (hollow markers).

\begin{figure}[H]
\centering
\includegraphics[width=0.6\columnwidth]{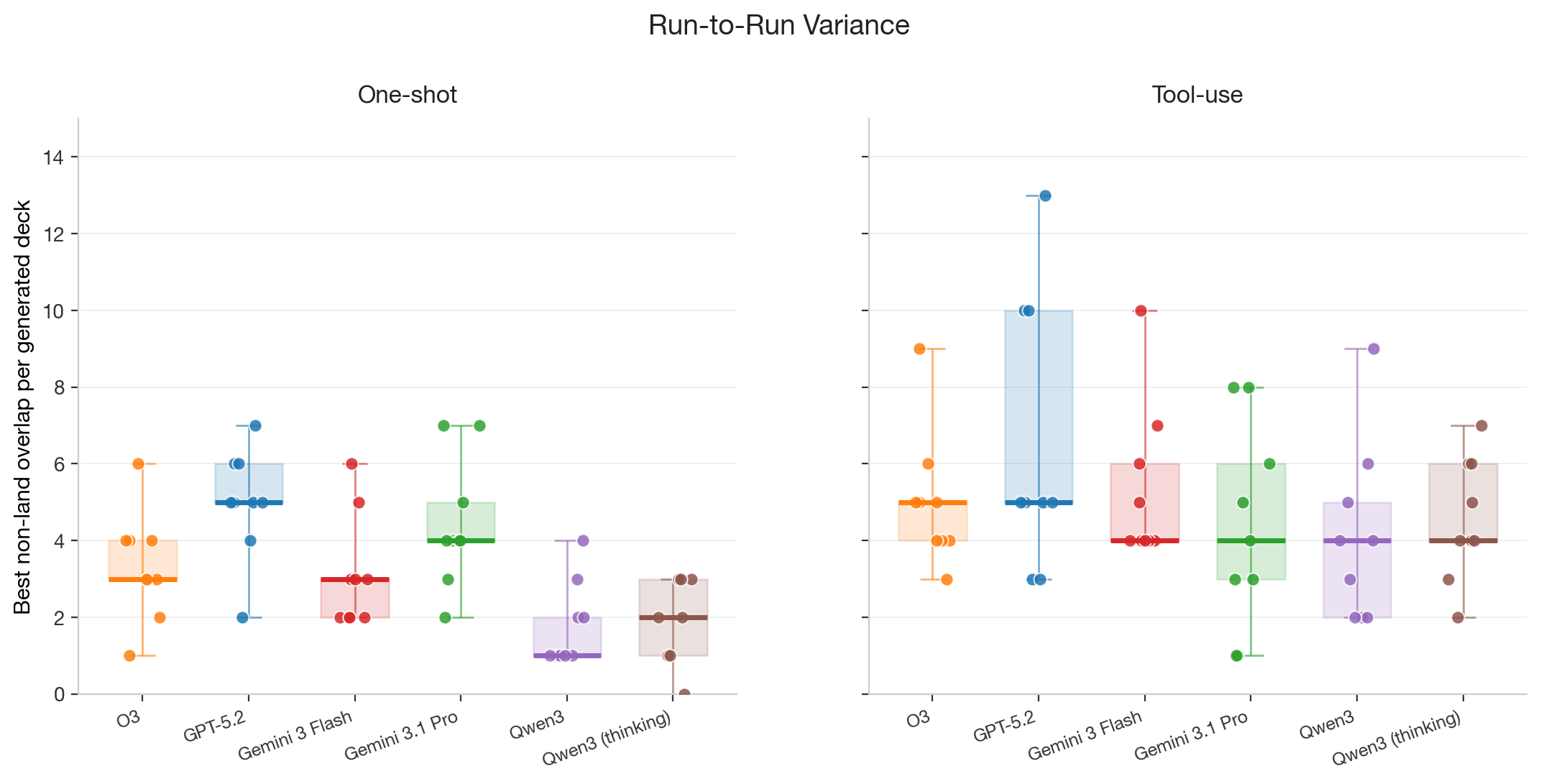}
\caption{Within-configuration spread of generated-deck quality. For each generated deck we record its maximum non-land card overlap with any PT deck. Each dot is one of the 9 decks per configuration. Boxes show the inter-quartile range, whiskers the min/max, and the central bar the median.}
\label{app:run-variance}
\end{figure}

Within-configuration spread is wider under tool-use than one-shot for every model, suggesting the scaffolding amplifies rather than dampens generation variance (\autoref{app:run-variance}). GPT-5.2 tool-use produces the single best non-land overlap (13) but also the widest inter-quartile range. Medians are otherwise comparable across configurations for most models, indicating that tool-use does not consistently improve the typical generated deck's structural similarity to a PT deck.

\begin{figure}[H]
\centering
\includegraphics[width=0.6\columnwidth]{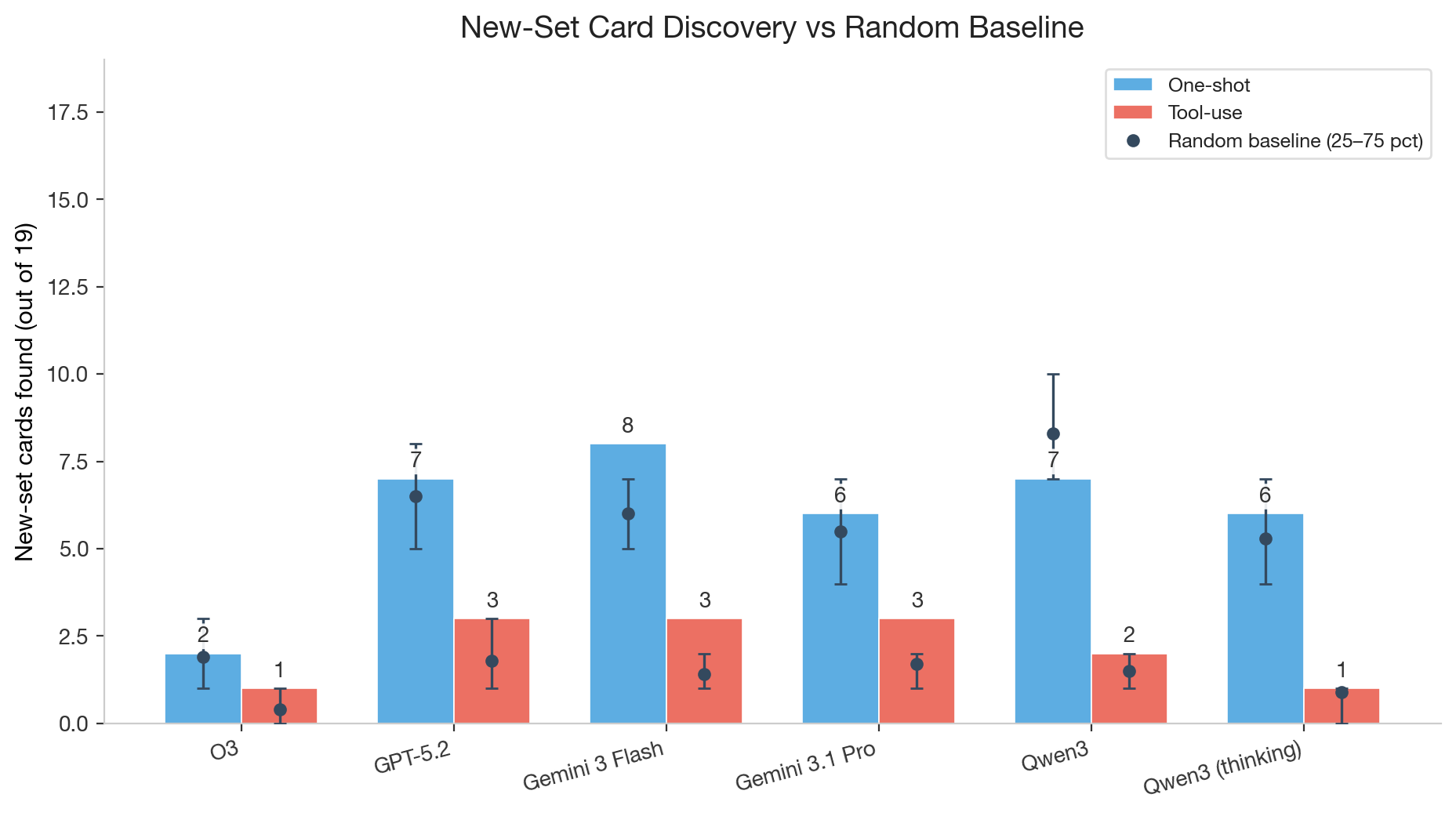}
\caption{Aggregate count of unique new cards found by each configuration (out of 19). Black whiskers overlay the configuration-matched random-baseline 25--75 percentile range estimated by Monte Carlo; a bar whose top exceeds the upper whisker sits above the 75th percentile of that configuration's random baseline. Gemini 3 Flash is the only model to do so in both configurations (8/19 one-shot, $1.34\times$ expected; 3/19 tool-use, $2.13\times$ expected).}
\label{fig:model-vs-random}
\end{figure}

\clearpage

\end{document}